\documentclass[letterpaper]{article}
\usepackage{aaai17}
\usepackage{times}
\usepackage{helvet}
\usepackage{courier}

\usepackage{algorithm}
\usepackage{algorithmic}

\usepackage{graphicx}
\usepackage{caption}
\usepackage{subcaption}
\graphicspath{ {images/} }

\usepackage{amssymb}
\usepackage{amsmath}
\let\emptyset\varnothing

\usepackage{tabu}
\usepackage{dblfloatfix}

\usepackage[hyphens]{url}

\title{Online Budgeted Learning for Classifier Induction}
\author{Eran Fainman \and Bracha Shapira \and Lior Rokach \and Yisroel Mirsky\\
Department of Software and Information Systems Engineering\\
Ben Gurion University of the Negev, Beer Sheva, Israel\\
e-mail: \{fainman,bshapira,liorrk,yisroel\}@bgu.ac.il
}
\begin{document}
\maketitle

\begin{abstract}
In real world machine learning applications, there is a cost associated with sampling different features. Budgeted learning can be used to select which feature-values to acquire from each instance in a dataset, such that the best model is induced under a given constraint. However, this approach is not possible in the domain of online learning since one may not retroactively acquire feature-values from past instances. In online learning, the challenge is to find the optimum set of features to be acquired from each instance upon arrival from a data stream. In this paper we introduce the issue of online budgeted learning and describe a general framework for addressing this challenge. We propose two types of feature value acquisition policies based on the multi-armed bandit problem: random and adaptive. Adaptive policies perform online adjustments according to new information coming from a data stream, while random policies are not sensitive to the information that arrives from the data stream. Our comparative study on five real world datasets indicates that adaptive policies outperform random policies for most budget limitations and datasets. Furthermore, we found that in some cases adaptive policies achieve near optimal results.
\end{abstract}

\section{Introduction and Motivation}
\noindent Predictive models are the core of many intelligent systems. A common approach to induce such models is to apply machine learning algorithms for classification. This type of algorithm trains classifiers based on historical data. The training data is represented as a set of instances, each described by a set of features and a discrete, dependent class label.

The features selected to describe a data instance may have a major effect on the performance of a learning algorithm. In many real world applications, the acquisition of these features can be expensive, and thus there is a trade-off between model performance and resource consumption. By intelligently selecting features, the training expense can be significantly reduced, while achieving an acceptable level of performance. Moreover, an intelligent selection of features results in a reduction in the need for data storage and processing, and contributes to a more efficient deployment of the learning algorithm. When data acquisition is limited by budget, intelligent feature acquisition becomes a necessity, and this domain is referred to as \textit{budgeted learning}.

Budgeted learning applies to supervised learning tasks where the class labels are fully specified but some or all of the feature-values are initially omitted. With budgeted learning, a learner sequentially acquires feature-values at a cost which may vary depending on the feature. The objective of the learner in this case (under an acquisition budget constraint), is to maximize the prediction performance by actively acquiring feature-values \cite{deng2013new}. 

Existing budgeted learning algorithms assume that it is possible to acquire all of an instance's feature-values throughout the training phase. However, this assumption does not always apply to real world situations where feature-values cannot be acquired at any time. In many real world scenarios, a learner must be trained in an online manner from data streams. A data stream is a potentially unbounded sequence of data instances. The challenge in this setting is that the feature-values of each current instance can only be acquired at the current time. For example, in the case of smartphone based activity recognition, for each time interval the learner must decide whether the location feature of the user should be acquired (via the smartphone's expensive GPS sensor) or to instead rely on other features for the task (e.g., device acceleration). Similar issues arise in a wide variety of real world situations involving tasks such as context prediction and health monitoring \cite{tayde2015context,perera2014context}.

In this paper we address budgeted learning in an online setting: given a data stream, a classifier-induction algorithm, the cost of acquiring each feature-value, and a fixed acquisition budget, the online budgeted learning issue is to identify the optimal set of features that should be acquired for each new instance, so as to maximize the induced classifier's performance, while adhering to the acquisition budget.

The contributions of this paper are (1) the creation of a general framework that addresses budgeted learning in an online setting, and (2) the establishment of four different feature value acquisition policies which can be used with the framework. We also describe an Oracle policy, which evaluates an upper bound of performance (note that the policy is not viable, since it utilizes instances from the future which are unavailable in a given current state). We evaluate the framework and acquisition policies using five real world datasets. The evaluation presents a trade-off analysis of the budget and the prediction accuracy of the resulting induced classifier. The proposed framework and policies are simple to implement and provide efficient online feature-value acquisition under different budget constraints.

\section{Related Work}
\noindent Previous work has addressed the issue of active feature-value acquisition \cite{saar2009active,bilgic2007voila,attenberg2011selective}. Active feature-value acquisition (AFA), which is similar to budgeted learning, acquires feature-values for missing data. The goal of AFA is to acquire feature-values in order to obtain a desired performance level at a minimal cost. The key difference between AFA and budgeted learning is that budgeted learning is limited by budget, while AFA does not have a strict budget. Like budgeted learning, AFA works under the assumption that feature-values can be acquired at any time during the training phase.

The feature-value acquisition problem is influenced by the classic multi-armed bandit problem, first introduced by Robbins (1952). The multi-arm bandit problem can be described as follows \cite{gittins1979bandit}: there are N arms, each having an unknown success probability of emitting a unit reward which are assumed to be independent of each other. The objective is to pull arms sequentially in order to maximize the total reward. Previous work by Deng et al. \cite{deng2013new} approached budgeted learning as a multi-armed bandit problem in which each pull is related to the acquisition of a feature-value for an instance, and the reward is related to the model's improved performance. We adopt the described approach but adjust it to the online setting.

An alternative line of research explores the domain of efficient feature-value acquisition from the perspective of \textit{adaptive sampling}. Many context-aware systems use the sensors of portable devices to build predictive models for classification. Since portable devices are very constrained in terms of power, the amount of sensing must be minimized. Previous research describes an architecture in which the sampling rates of sensors are adapted according to the information they provide about the possible next contexts \cite{wood2012adaptive}. This approach assumes that instances are time-dependent, so the context transition can be represented as a Markov chain. In our work, we assume that the order in which the data stream produces instances may not be relevant, and thus cannot be temporally modeled.

\section{Problem Formulation and Framework}
\noindent Given the challenges of real world settings described above, our goal is to place an online budgeted feature-value acquisition framework between a data stream $S$ and a given classifier $C$. Instances are produced by $S$ in a sequential manner. An instance $x$ is represented as a set of features $F=\{f_1,...,f_n\}$ and a discrete, dependent class label $y$. When a new instance arrives from $S$, both the feature-values and the class label value are initially unknown. The feature-value for $f_i$ of instance $x$ can be acquired at a given fixed cost $f^c_i$. Let $AF(x)$ be a subset of $F$ which represents the set of acquired features for an instance $x$. We denote the total acquisition cost for instance $x$ as $c(x)$ where 
\begin{equation}
c(x)=\sum\limits_{f_i \in AF(x)}^{} {f^c_i}
\end{equation}
Similar to the classic budgeted learning problem, we assume that there is no cost associated with acquiring the value of the class label $y$. Let $B \in \mathbb{R}$ be the total budget constraint. The budget for feature-value acquisition of an instance $x$ is denoted as $b(x)$. Post-acquisition instances are stored in the training set $T$. Finally, the total budget $B$ fulfills: $\sum_{x \in T}^{} {b(x)} \leq B$. An acquisition policy $\pi$ allocates budget $b(x)$ for each instance $x$ and seeks the optimal $AF(x)$ which fulfills $c(x) \leq b(x)$. The objective of $\pi$ is to maximize the performance of classifier $C$ which is induced based on the acquired training set $T$.

The framework follows a generic iterative procedure (pseudo code available in Algorithm \ref{alg1}):
\begin{enumerate}
\item New instance $x$ arrives from $S$.
\item Policy $\pi$ determines $b(x)$.
\item Policy $\pi$ iteratively acquires feature-values for $x$.
\item The class label for $x$ is sampled.
\item The instance $x$ is passed to the training set $T$. 
\end{enumerate}

In order to ensure that features that exceed the budget are not selected, the framework maintains a so-called "potential features set", denoted as $PF(x)$. The $PF(x)$ is initialized with the $F$ for each new upcoming instance $x$. After each feature-value acquisition iteration, features that exceed the remaining budget are removed from the set.

Our suggested framework is based on the following assumptions: (1) the decision whether to acquire a feature-value can be made only when the instance is active, e.g., if we collect data from a mobile phone, and we decided not to collect data from a specific sensor at a particular time, we can't go back in time and acquire it. (2) The sequential feature-value acquisition for an instance is supported by assuming negligible acquisition and decision times. (3) The class label value for an instance is provided at the end of the feature-value acquisition phase, so it has no effect on the decision regarding which feature-values to acquire. (4) Misclassification costs are considered to be equal. (5) For simplification, the data stream length is assumed to be predefined.

\begin{algorithm}[h]
\caption{General Online Budgeted Feature-Value Acquisition Framework}
\label{alg1}
\begin{algorithmic}[1]
\REQUIRE
\item[]
$S$ - sequential stream of instances

$B$ - acquisition budget

$\{f_1^c,...,f_n^c\}$ - feature-value acquisition cost vector

\ENSURE
\item[]
$T$ - set of training instances

\STATE Initialize $T \leftarrow \emptyset$
\STATE Initialize $\pi$
\WHILE {a new instance $x$ arrives from $S$} \label{newinst}
\STATE Initialize acquisition cost $c(x)=0$
\STATE Initialize $AF(x) \leftarrow \emptyset$
\STATE Initialize budget for instance $b(x)$ according to $\pi$
\STATE Initialize feature set $PF(x) \leftarrow \{\forall f_i \in F | f_i^c \leq b(x)\}$
\WHILE {$PF(x)$ is not empty}
\STATE Select feature $f_{best}$ from $PF(x)$ according to $\pi$  \label{ftracqiterstart}
\STATE Acquire value for $f_{best}$
\STATE Update $c(x) \leftarrow c(x)+f_{best}^c$
\STATE Update $AF(x) \leftarrow AF(x) \cup f_{best}$
\STATE Remove $f_{best}$ from $PF(x)$
\STATE Update $PF(x) \leftarrow \{\forall f_i \in PF(x) | f_i^c + c(x) \leq b(x)\}$ \label{ftracqiterend}
\ENDWHILE
\STATE Sample value for class label $y$
\STATE Update training set $T \leftarrow T \cup x$
\STATE Update policy $\pi$
\ENDWHILE
\RETURN $T$
\end{algorithmic}
\end{algorithm}

\section{Acquisition Policies}
\noindent In this section, we describe the four acquisition policies which we implement for our framework. The four polices reflect the multi-armed bandit problem and can be categorized as random policies and adaptive policies. The two policy types differ from each other by (1) the information on which they base their decisions, (2) the method for the division of the budget between the instances, and (3) the way that new information from the stream is utilized.

\subsection{Random Policies}
We define random policies as policies which are not sensitive to the information that arrives from the data stream. Random policies operate in a stochastic manner, where each feature has a probability to be selected for acquisition. The probabilities are generated at the start of each feature-value acquisition iteration (see line \ref{ftracqiterstart} in Algorithm \ref{alg1}). Only prior knowledge, e.g., acquisition costs, may be taken into account when probabilities are generated. The budget for each instance is the same and is calculated as $b(x)=\frac{B}{|S|}$.

\subsubsection{Pure Random Acquisition Policy} The pure random acquisition policy selects features following a uniform random distribution. Specifically, the probability to select $f_i$ from $PF(x)$ is
\begin{equation}
p(f_i) = \frac{1}{|PF(x)|}
\end{equation}
 We use this policy as the baseline for our evaluation.

\subsubsection{Cost-Sensitive Random Acquisition Policy} In the cost-sensitive random acquisition policy, features are selected randomly according to a probability that is negatively correlated with the feature's acquisition cost. Specifically, the probability to select $f_i$ from $PF(x)$ is
\begin{equation}
p(f_i) = \frac{\frac{1}{f^c_i}}{\sum\limits_{f_j \in PF(x)}^{} \frac{1}{f^c_j}}
\end{equation}
The motivation for the use of this policy is that we want to acquire features with low acquisition cost more often than other features.

\subsection{Adaptive Policies} \label{adaptivesection}
\noindent A policy is defined as an adaptive policy if it utilizes new information coming from the data stream. Adaptive policies approach the problem of online budgeted learning as an \textit{explore} or \textit{exploit} problem. When deciding which feature to select next, the policy decides whether to exploit the features that have been considered as valuable for previous instances, or to explore the other features. Our suggested adaptive policies tend to favor exploitation, therefore we involve stochastic elements in order to facilitate exploration.

\begin{figure}[h]
\centering
\includegraphics[width=\columnwidth]{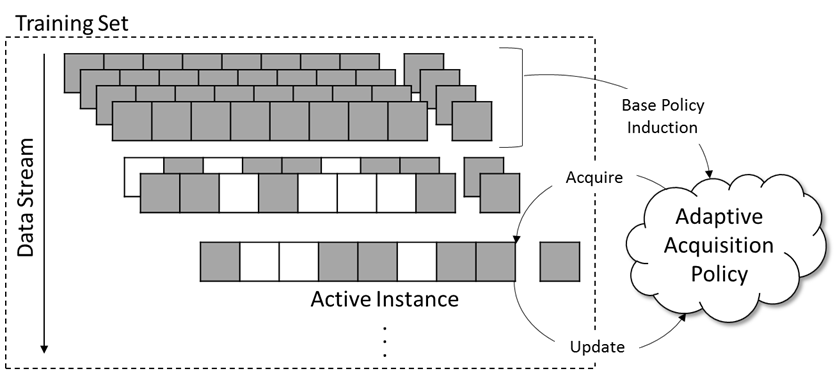}
\caption{Adaptive acquisition policy illustration}
\label{adaptive}
\end{figure}

The adaptive policies require initial data upon which to base their exploitation. Therefore, we allocate part of the total acquisition budget for complete feature-value acquisition of the first instances. We denote the set of complete instances as $\widehat{S}$. Then, we induce a base acquisition policy that relies on $\widehat{S}$. The budget for each of the remaining instances is calculated as $b(x)=\frac{\widetilde{B}}{|\widetilde{S}|}$ where the rest of the budget and the stream of the remaining instances are denoted as $\widetilde{B}$ and $\widetilde{S}$, respectively. The acquisition policy is updated when a new instance is added to the training set. An illustration of the process is presented in Figure \ref{adaptive}.

\subsubsection{Variance Cost-Sensitive Acquisition Policy} In the variance cost-sensitive acquisition policy, we first acquire complete instances as previously described. Then, we calculate the variance of each of the features. For the rest of the instances, features are selected following a probability which is positively correlated with the feature's variance divided by the feature's cost. Before calculating a feature's variance we rescale the feature's observed values to the range [0,1] using min-max normalization to avoid unit bias. Let $f^*_i$ be the rescaled version of $f_i$. The variance of $f_i$ is denoted as $f^v_i$ and defined as
\begin{equation}
f^v_i = \frac{\sum\limits_{x \in T}{} {(f^*_i(x)-\bar{f^*_i})^2}}{|T|-1}
\end{equation}
The probability to select $f_i$ from $PF(x)$ is defined as
\begin{equation}
p(f_i) = \frac{\frac{f^v_i}{f^c_i}}{\sum\limits_{f_j \in PF(x)}^{} {\frac{f^v_j}{f^c_j}}}
\end{equation}
With each feature-value acquisition, the feature's respective variance is recalculated.

This policy is based on the notion that features with high variance may describe the dependent variable better than other features. This idea is inspired by the principal component analysis (PCA) algorithm which seeks the principal component that has the greatest possible variance. We divide the variance by the acquisition cost in order to prioritize the less expensive features. Thus, we would like to acquire features with greater variance and lower acquisition cost more often than other features.

\subsubsection{Tree Classifier Based Acquisition Policy}
In the tree classifier based acquisition policy, we first acquire complete instances as previously described. With these instances we induce a tree based classifier using a cost-sensitive version of the C4.5 algorithm \cite{lauwereins2014context}. The decision tree, described by Lauwereins et al., selects features for tree splits according to their trade-off between information gain and power consumption (in our case, the power consumption is redefined as feature acquisition cost). When a new instance arrives from the data stream, we perform a walk through the tree from the root node down. The inner nodes visited during the walk indicate useful features. Therefore, we select and acquire those features until (1) we select a feature that is not contained in the $PF(x)$, or (2) we reach a leaf node. In either case the rest of the features in the $PF(x)$ are acquired according to the variance cost-sensitive acquisition policy. Once the budget has been met and the instance has been added to the training set $T$, the decision tree is rebuilt with the latest training data.

This policy is motivated by the idea that decision trees represent a collection of hypotheses. Therefore, when a new instance arrives, its matching hypothesis, represented by a path in the tree, is tested through the instance. Thus, exploitation is applied. Moreover, the policy encourages exploration, since it may select features that are not covered by the matching hypothesis.

\section{Experiments}
\noindent We tested the framework and acquisition policies on five real world datasets. The datasets reflect real world domains where the training data is collected in an online manner and a cost can be associated with the acquisition of each feature-value. Four of the datasets were obtained from the UCI repository \cite{Lichman:2013,reyes2016transition}. The fifth dataset, a smartphone sensor dataset involving SMS malware, was collected in our labs.

The reason we collected the fifth dataset is to illustrate a real world scenario in which online budgeted learning is a necessity. With the high adoption of smart mobile devices among world's population, the role of machine learning on these devices becomes important (e.g., see \cite{mirsky2016anomaly}). However, most of the smart mobile devices, e.g., smartphones, are limited in storage and power resources. We demonstrated the contribution of online budgeted learning by using this dataset to detect malicious SMSs. The dataset contains features which capture the motion of a user's smartphone during the transmission of SMSs over the course of several months. In order to compile the dataset, we placed a malicious agent on the user's smartphone; the agent sends SMSs simulating the behavior of actual sophisticated SMS malware aimed at financial theft (a malware that sends SMSs to a premium phone number in order to charge the user's account). Each instance in the dataset captures the transmission of an SMS with 149 features describing the smartphone's acceleration, rotation, orientation, and location. A label is provided to each instance to indicate whether the transmission was malicious or benign.

Table \ref{meta} provides a detailed comparison of the five datasets. The datasets can be divided into two dataset collections: (1) a collection in which mobile device sensors are sampled, denoted as the \textit{sensors} collection. This collection consists the HAPT dataset - a smartphone-based recognition of human activities and postural transitions dataset \cite{reyes2016transition}, and our own collected SMS leakage dataset. (2) A collection in which medical tests are performed, denoted as the \textit{medical} collection. This collection consists the Cardiotocography and the Thyroid datasets \cite{Lichman:2013}. The sensors collection datasets are characterized by a large number of features and instances, while the medical collection datasets are characterized by significantly lower dimensions. All of the datasets have no missing values, so the acquisition policies were not restricted. For most of the datasets, feature acquisition costs were not provided. Therefore, for the sensors collection datasets we gave each feature an acquisition cost which is associated with its respective sensor power consumption. Some of the sensor power consumption data was obtained from previous research by "Google",\footnote{\path{https://dl.google.com/io/2009/pres/W_0300_CodingforLife-BatteryLifeThatIs.pdf}} while the rest were randomly generated and fine-tuned based on our prior knowledge. The sensor power consumption data is presented in Table \ref{costs}. For the medical collection datasets (except the Thyroid dataset), acquisition costs were randomly generated.

\begin{table}[h]
\centering
\resizebox{\columnwidth}{!}{%
\begin{tabular}{c c c c c}
\hline
Dataset & Collection & \#Instances & \#Attributes & \#Classes \\
\hline
\hline
HAPT & Sensors & 10,929 & 561 & 12 \\
SMS Leakage (SMSL) & Sensors & 90,528 & 149 & 2 \\
Cardiotocography-3 (CTG3) & Medical & 2,128 & 21 & 3 \\
Cardiotocography-10 (CTG10) & Medical & 2,128 & 21 & 10 \\
Thyroid & Medical & 7,200 & 21 & 3 \\
\hline
\end{tabular}%
}
\caption{\label{meta}Dataset information}
\end{table}

\begin{table}[h]
\centering
\resizebox{\columnwidth}{!}{%
\begin{tabular}{c c c c c}
\hline
Sensor & Evaluated Power Consumption (mA) \\
\hline
\hline
Accelerometer & 10 \\
Gyroscope & 9 \\
Location & 140 \\
Orientation & 10 \\
Rotation & 12 \\
\hline
\end{tabular}%
}
\caption{\label{costs} Sensor Power Consumption Data}
\end{table}

We evaluated the four acquisition policies previously described on each of the datasets. Each acquisition policy was tested with nine different total acquisition budget values. Each acquisition budget was calculated with the percent $\alpha$ of the maximum cost $|S|*\sum_{f_i \in F}{} {f^c_i}$, where $\alpha \in \{0.1,0.2,0.3,0.4,0.5,0.6,0.7,0.8,0.9\}$. The performance of each acquisition policy was evaluated over 10 independent runs. In each run, 70\% of the instances were randomly selected and treated as the data stream. The test set was formed by the remaining instances. In each run, when the acquisition phase was completed, we trained an XgBoost classifier \cite{chen2016xgboost} on the training set. The XgBoost classifier was initialized with fixed parameters for each of the datasets. Parameters were configured based on best practice heuristics.\footnote{\path{https://www.analyticsvidhya.com/blog/2016/03/complete-guide-parameter-tuning-xgboost-with-codes-python}} The prediction accuracy of the test set (measured by multi-class area under the ROC curve (AUC)) served as the performance measure of the induced classifier. For the adaptive policies, we used 20\% of the budget for acquiring the complete instances, and the rest of the budget was used to acquire features dynamically as described in section \ref{adaptivesection}. We used this portion of the budget based on preliminary experiments that we conducted to determine the optimal ratio to use. Since there is no difference between the policies when there are no budget constraints, we performed complete feature-value acquisition for each of datasets within each run described above.

As it is not feasible to determine an optimal acquisition policy to compare our proposed methods to, we needed a simple method for estimating the upper bound of performance. Therefore, we defined an Oracle acquisition policy which assumes that a given Oracle provides a list of features, ordered based on their trade-off between information gain and cost. The policy allocates equal budget for each instance and selects features according to their order in the list. Although the Oracle acquisition policy is not a viable acquisition approach, it does provide a simple estimation of the upper bound of performance. For our experiments, an ordered list of features with their information gain was generated by training the XgBoost classifier with the complete data.

\begin{figure*}[b]
\centering
\begin{subfigure}[b]{\columnwidth}
\includegraphics[width=\columnwidth]{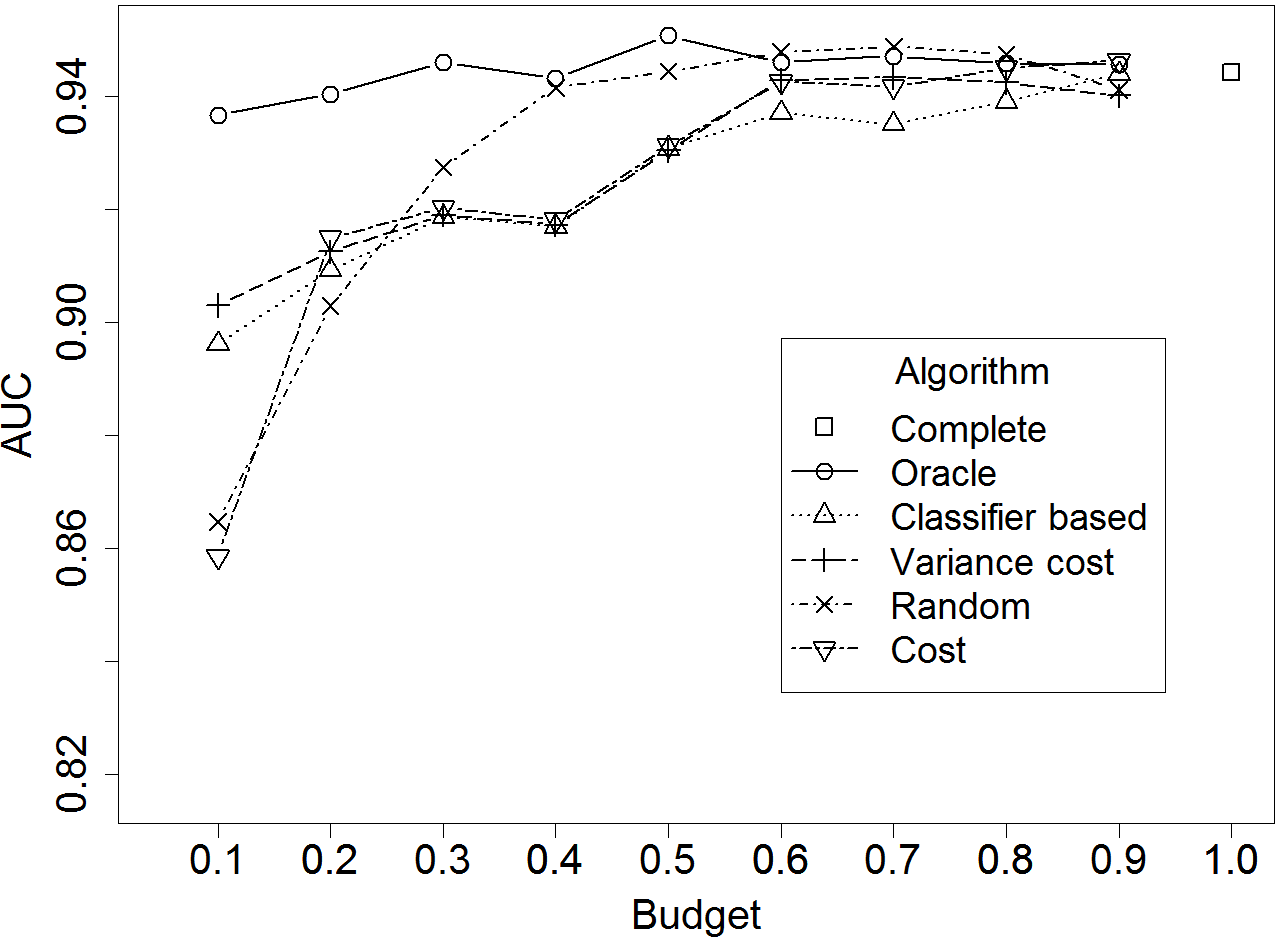}
\caption{HAPT}
\label{fig:HAPT}
\end{subfigure}
\qquad
\begin{subfigure}[b]{\columnwidth}
\includegraphics[width=\columnwidth]{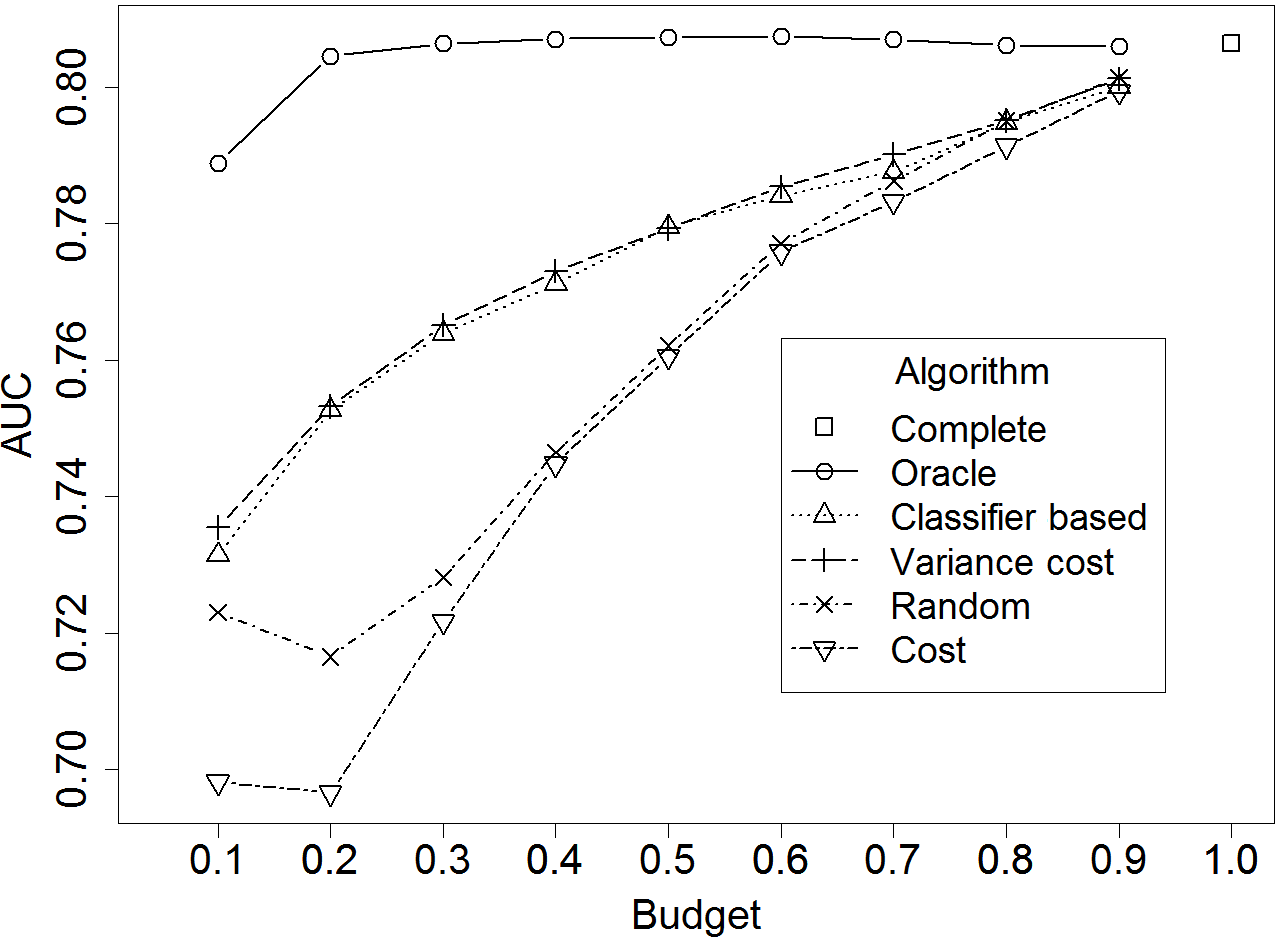}
\caption{SMSL}
\label{fig:SMSL}
\end{subfigure}
\vspace{8pt}

\begin{subfigure}[b]{\columnwidth}
\includegraphics[width=\columnwidth]{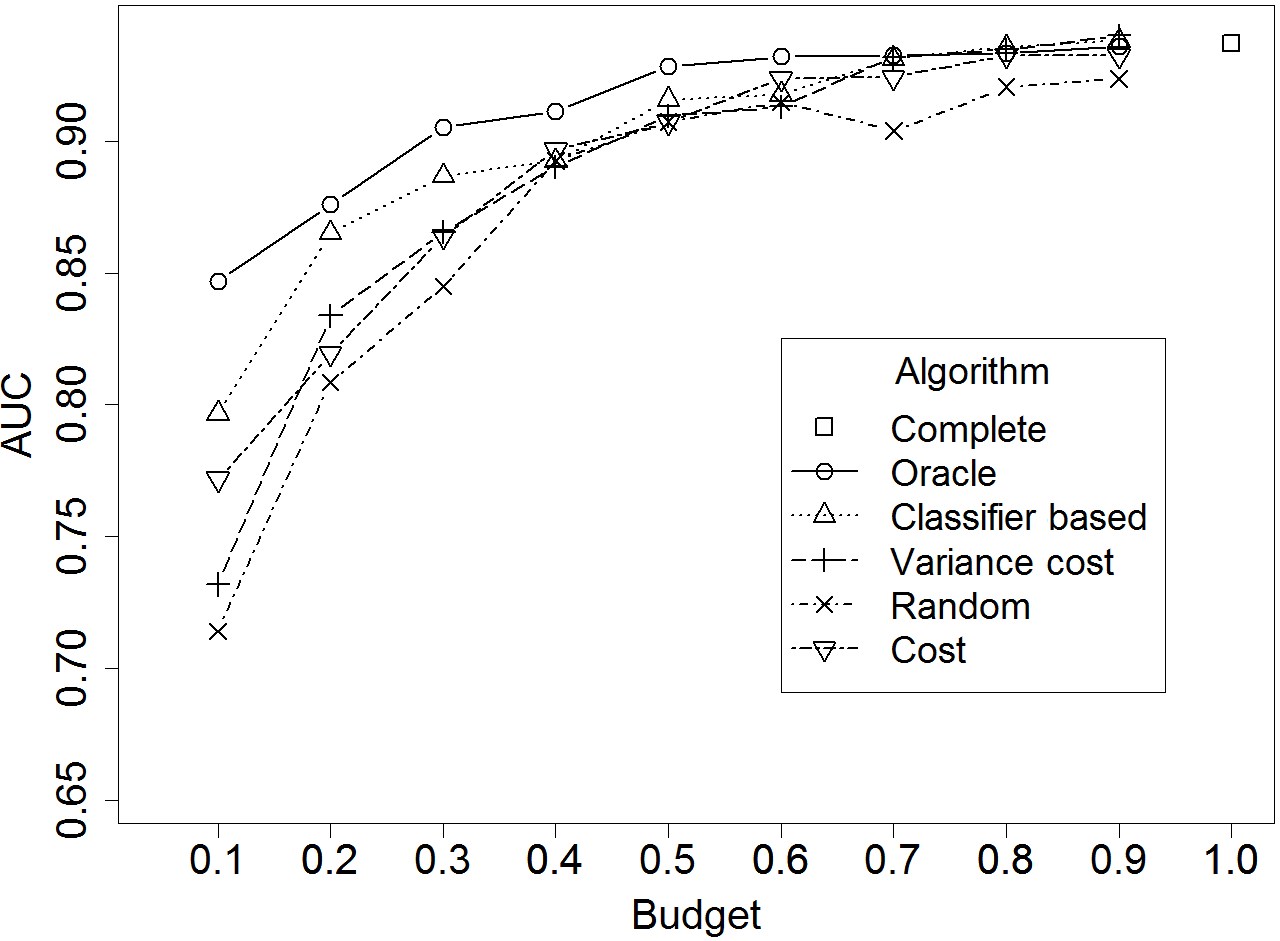}
\caption{CTG3}
\label{fig:CTG3}
\end{subfigure}
\qquad
\begin{subfigure}[b]{\columnwidth}
\includegraphics[width=\columnwidth]{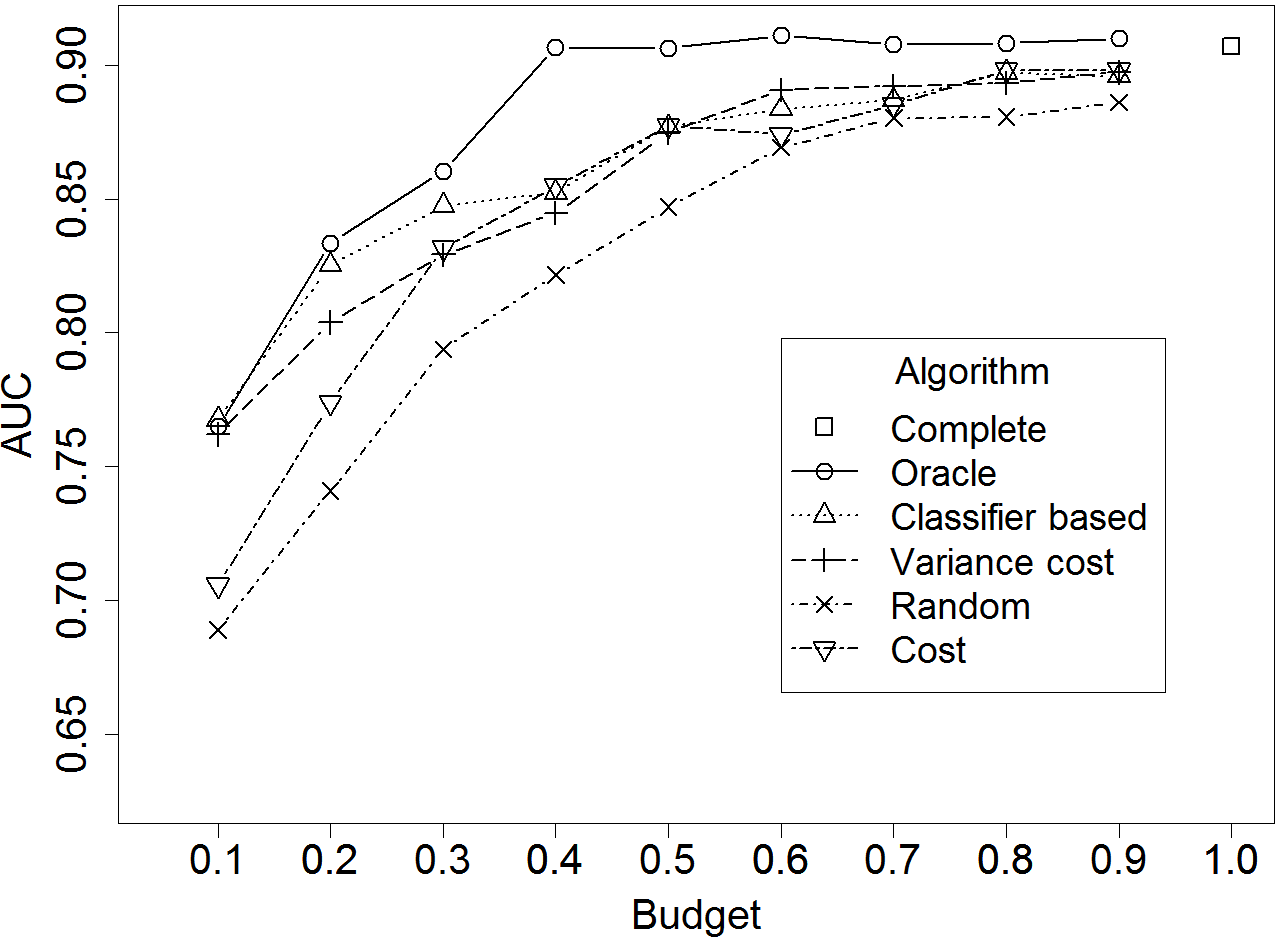}
\caption{CTG10}
\label{fig:CTG10}
\end{subfigure}
\vspace{8pt}

\begin{subfigure}[b]{\columnwidth}
\includegraphics[width=\columnwidth]{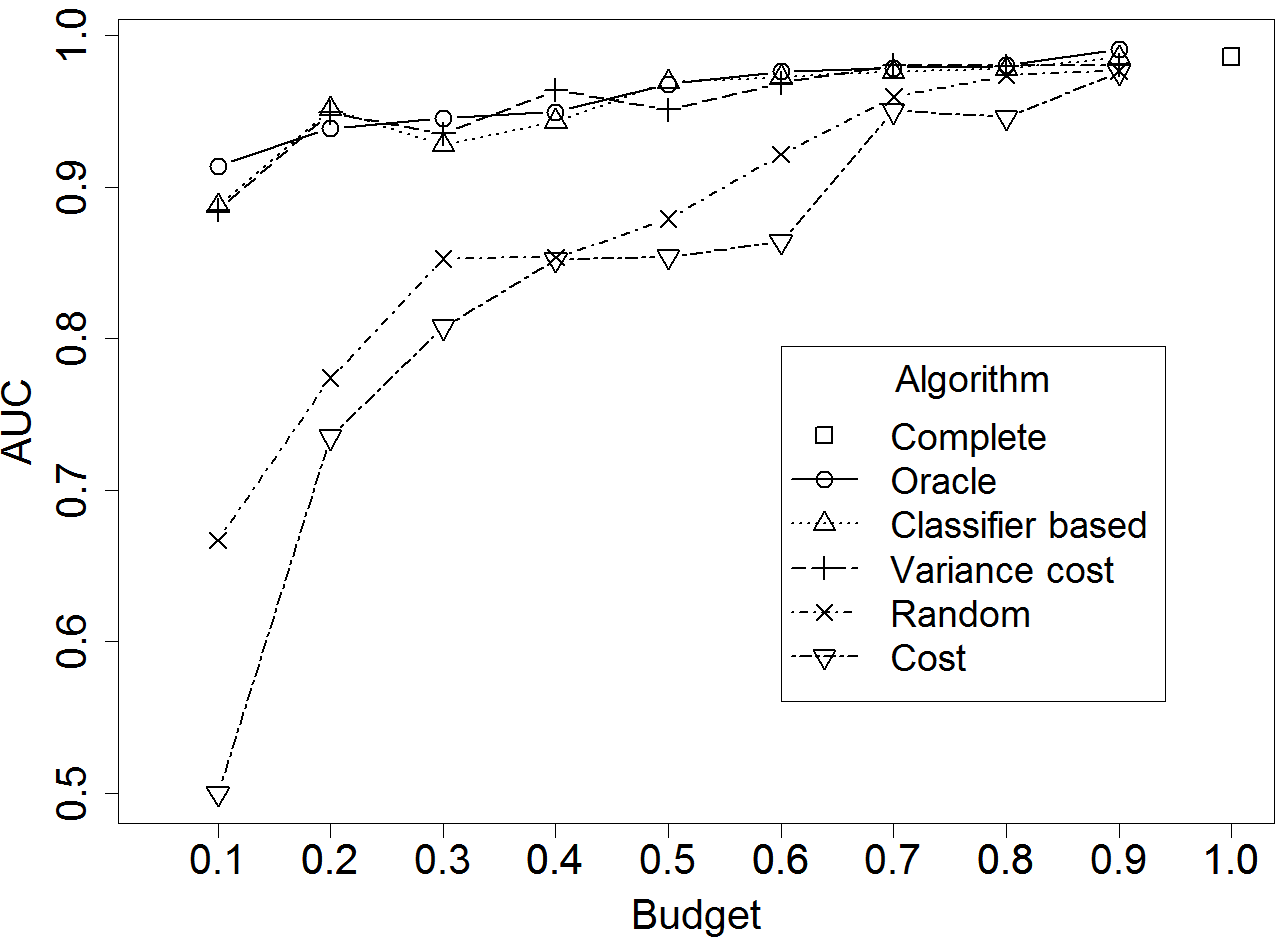}
\caption{Thyroid}
\label{fig:Thyroid}
\end{subfigure}
\caption{Comparing Acquisition Policies Using Different Budgets and Data Sets}
\label{results}
\end{figure*}

To evaluate the overall behavior of each of the acquisition policies, we constructed curves that reflect the performance of the induced classifier for each of the budget constraints. Each point in the graphs represents the mean of the area under the ROC curve (AUC) over the 10 runs for the specified budget and acquisition policy. To keep the plots uncluttered, in figure \ref{results} we plot results for the complete data acquisition (Complete), the random acquisition policy (Random), the cost-sensitive acquisition policy (Cost), the variance cost-sensitive acquisition policy (Variance cost), the tree classifier based acquisition policy (Classifier based) and the Oracle acquisition policy (Oracle).

As can be seen in figure \ref{results}, as expected, the Oracle policy achieved the best performance for all of the datasets and $\alpha$ values, however as mentioned previously, it is not a viable approach. The adaptive policies performed better than the random policies, except for the HAPT dataset. For the HAPT dataset, the random policies performed equal to or better than the adaptive policies for $\alpha \geq 0.3$. Our explanation for this is that since the HAPT dataset has the largest number of features, the adaptive policies tend to exploit a small set of its features, while other more promising features are not explored. For the Thyroid dataset, the adaptive policies obtained similar results to those achieved by the Oracle policy. For both the SMS malware and Thyroid datasets, the adaptive policies outperformed the random policies. The difference was more significant with small $\alpha$ values. Our explanation for the latter is that when there is a small budget and a large set of features, identifying the optimal set of features becomes a harder task. Therefore, we expect that good policies will obtain significant better performance in such cases. With regards to the adaptive policies, we found that the variance cost-sensitive policy outperformed the tree classifier based policy with the sensor dataset collection. However, with the medical dataset collection, the two policies obtained similar results. We believe these results are because the variance cost-sensitive policy is more exploratory than the tree classifier based policy.

\section{Conclusions and Future Work}
\noindent We presented a general framework to address online budgeted feature-value acquisition, in which instances arrive from a data stream and an acquisition policy selects which features should be acquired within given budget constraints. We also introduced four acquisition policies categorized as random or adaptive policies and defined an Oracle policy, which is not viable for acquisition but provides an upper bound of performance. We evaluated the acquisition policies on five real world datasets with various budget constraints and demonstrated that the adaptive acquisition policies outperform the random policies for most of the budget values and datasets. In some cases, the adaptive acquisition policies achieved similar results to those achieved by the Oracle policy.

There are several directions for future work. First, based on our observations of the large datasets, it would be beneficial to be able to control the amount of exploration and exploitation an adaptive policy should perform. A possible direction for the latter may be influenced by the simulated annealing technique \cite{aarts1988simulated}. Furthermore, we plan to test the computational resources (e.g., machine time) required to train the classifier based on the training data acquired under different budget constraints. Moreover, we may develop acquisition methods for problems in which the length of the data stream is unknown. To sum up, we believe that future research in this area is necessary for many real world applications in which feature-value acquisition is limited by budget constraints and must be performed in an online manner.

\clearpage
\bibliographystyle{aaai}
\bibliography{bibliography}

\begin{thebibliography}{}

\bibitem[\protect\citeauthoryear{Aarts and Korst}{1988}]{aarts1988simulated}
Aarts, E., and Korst, J.
\newblock 1988.
\newblock Simulated annealing and boltzmann machines.

\bibitem[\protect\citeauthoryear{Attenberg \bgroup et al\mbox.\egroup
  }{2011}]{attenberg2011selective}
Attenberg, J.; Melville, P.; Provost, F.; and Saar-Tsechansky, M.
\newblock 2011.
\newblock Selective data acquisition for machine learning.
\newblock {\em Cost-sensitive machine learning}.

\bibitem[\protect\citeauthoryear{Bilgic and Getoor}{2007}]{bilgic2007voila}
Bilgic, M., and Getoor, L.
\newblock 2007.
\newblock Voila: Efficient feature-value acquisition for classification.
\newblock In {\em PROCEEDINGS OF THE NATIONAL CONFERENCE ON ARTIFICIAL
  INTELLIGENCE}, volume~22,  1225.
\newblock Menlo Park, CA; Cambridge, MA; London; AAAI Press; MIT Press; 1999.

\bibitem[\protect\citeauthoryear{Chen and Guestrin}{2016}]{chen2016xgboost}
Chen, T., and Guestrin, C.
\newblock 2016.
\newblock Xgboost: A scalable tree boosting system.
\newblock {\em arXiv preprint arXiv:1603.02754}.

\bibitem[\protect\citeauthoryear{Deng \bgroup et al\mbox.\egroup
  }{2013}]{deng2013new}
Deng, K.; Zheng, Y.; Bourke, C.; Scott, S.; and Masciale, J.
\newblock 2013.
\newblock New algorithms for budgeted learning.
\newblock {\em Machine learning} 90(1):59--90.

\bibitem[\protect\citeauthoryear{Gittins}{1979}]{gittins1979bandit}
Gittins, J.~C.
\newblock 1979.
\newblock Bandit processes and dynamic allocation indices.
\newblock {\em Journal of the Royal Statistical Society. Series B
  (Methodological)}  148--177.

\bibitem[\protect\citeauthoryear{Lauwereins \bgroup et al\mbox.\egroup
  }{2014}]{lauwereins2014context}
Lauwereins, S.; Badami, K.; Meert, W.; and Verhelst, M.
\newblock 2014.
\newblock Context-and cost-aware feature selection in ultra-low-power sensor
  interfaces.
\newblock In {\em European Symposium on Artificial Neural Networks,
  Computational Intelligence and Machine Learning},  93--98.

\bibitem[\protect\citeauthoryear{Lichman}{2013}]{Lichman:2013}
Lichman, M.
\newblock 2013.
\newblock {UCI} machine learning repository.

\bibitem[\protect\citeauthoryear{Mirsky \bgroup et al\mbox.\egroup
  }{2016}]{mirsky2016anomaly}
Mirsky, Y.; Shabtai, A.; Shapira, B.; Elovici, Y.; and Rokach, L.
\newblock 2016.
\newblock Anomaly detection for smartphone data streams.
\newblock {\em Pervasive and Mobile Computing}.

\bibitem[\protect\citeauthoryear{Perera \bgroup et al\mbox.\egroup
  }{2014}]{perera2014context}
Perera, C.; Zaslavsky, A.; Christen, P.; and Georgakopoulos, D.
\newblock 2014.
\newblock Context aware computing for the internet of things: A survey.
\newblock {\em IEEE Communications Surveys \& Tutorials} 16(1):414--454.

\bibitem[\protect\citeauthoryear{Reyes-Ortiz \bgroup et al\mbox.\egroup
  }{2016}]{reyes2016transition}
Reyes-Ortiz, J.-L.; Oneto, L.; Sam{\`a}, A.; Parra, X.; and Anguita, D.
\newblock 2016.
\newblock Transition-aware human activity recognition using smartphones.
\newblock {\em Neurocomputing} 171:754--767.

\bibitem[\protect\citeauthoryear{Saar-Tsechansky, Melville, and
  Provost}{2009}]{saar2009active}
Saar-Tsechansky, M.; Melville, P.; and Provost, F.
\newblock 2009.
\newblock Active feature-value acquisition.
\newblock {\em Management Science} 55(4):664--684.

\bibitem[\protect\citeauthoryear{Tayde and Bhala}{2015}]{tayde2015context}
Tayde, K., and Bhala, A.
\newblock 2015.
\newblock Context awareness in mobile computing.
\newblock {\em International Journal} 3(7).

\bibitem[\protect\citeauthoryear{Wood \bgroup et al\mbox.\egroup
  }{2012}]{wood2012adaptive}
Wood, A.~L.; Merrett, G.~V.; Gunn, S.~R.; Al-Hashimi, B.~M.; Shadbolt, N.~R.;
  and Hall, W.
\newblock 2012.
\newblock Adaptive sampling in context-aware systems: a machine learning
  approach.
\newblock In {\em Wireless Sensor Systems (WSS 2012), IET Conference on},
  1--5.
\newblock IET.

\end{thebibliography}

\end{document}